\documentclass[10pt,twocolumn,letterpaper]{article}

\usepackage{cvpr}              
\usepackage{multirow}
\usepackage[table]{xcolor}
\usepackage{arydshln}
\usepackage{wrapfig}
\usepackage[dvipsnames, svgnames, x11names]{xcolor}
\definecolor{RLHFcolor}{HTML}{709DE5}
\definecolor{RLVFcolor}{HTML}{CD3E37}
\definecolor{RLSFcolor}{HTML}{EC802D}
\definecolor{RLDFcolor}{HTML}{53AFA8}
\definecolor{Best}{HTML}{F5B7BF}
\definecolor{Secondbest}{HTML}{D5DEFF}
\definecolor{CVPR}{HTML}{8080FF}
\usepackage{tcolorbox}
\definecolor{cvprblue}{rgb}{0.21,0.49,0.74}
\usepackage[pagebackref,breaklinks,colorlinks,allcolors=cvprblue]{hyperref}
\usepackage{graphicx} 

\def\paperID{ } 
\def\confName{CVPR}
\def\confYear{2026}

\title{\raisebox{-1ex}{\includegraphics[height=4ex]{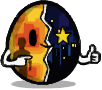}} Fusion in Your Way: Aligning Image Fusion with Heterogeneous Demands via Direct Preference Optimization}
\author{Weijian Su$^{1,2}$\quad Songqian Zhang$^{1,2}$\quad Yuqi Han$^{1,2*}$\quad Jian Zhuang$^{1,2}$\quad Yongdong Huang$^3$\quad Qiang Zhang$^{1,2*}$\\
{\small$^1$ School of Computer Science and Technology, Dalian University of Technology}\\
{\small$^2$ Key Laboratory of Social Computing and Cognitive Intelligence (Dalian University of Technology), Ministry of Education}\\
{\small$^3$ Institute of Image Processing and Understanding, North Minzu University}\\
{\tt\small suwj@mail.dlut.edu.cn \quad yqhanSCST@dlut.edu.cn \quad zhangq@dlut.edu.cn}
}

\begin{document}
\twocolumn[{%
\renewcommand\twocolumn[1][]{#1}%
\maketitle
\includegraphics[width=1\linewidth]{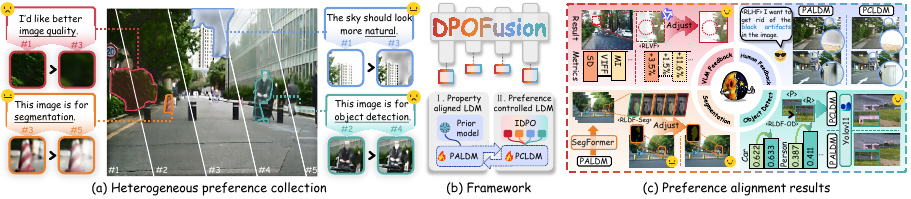}
\captionof{figure}{Overview of our DPOFusion framework. (a) We collect heterogeneous preference demands for a given scene from diverse evaluators. (b) The DPOFusion framework consists of a property-aligned latent diffusion model and a preference-controlled latent diffusion model. (c) Our method achieves precise preference alignment, fine-tuning the fused results for various human and machine-vision tasks.}
\label{fig:teaser}
}]

\let\thefootnote\relax\footnotetext{\hspace{-1em} *Corresponding authors.}
\setlength{\dbltextfloatsep}{3em}
\begin{abstract}
As a key technique in multi-modal processing, infrared and visible image fusion (IVIF) plays a crucial role in integrating complementary spectral information for visual enhancement and downstream vision tasks. Despite remarkable progress, existing methods struggle to flexibly accommodate heterogeneous demands. Achieving adaptive fusion that aligns with various preferences from both human and machine vision remains an open and challenging problem.
To address this challenge, we propose DPOFusion, a direct preference optimization (DPO) framework integrating the property-aligned latent diffusion model (PALDM) and the preference-controllable latent diffusion model (PCLDM), enabling task-guided, preference-adaptive IVIF for both human and machine vision. The PALDM leverages a latent fusion prior and a joint conditional loss to generate diverse candidate fusion results with various properties. PCLDM is subsequently fine-tuned via instance direct preference optimization (IDPO), enabling direct control of the final fusion results with heterogeneous preference signals.
Experimental results demonstrate that our framework not only attains precise preference alignment among humans, vision-language models, and task-driven networks, but also sets a new benchmark for adaptive fusion quality and task-oriented transferability.  Our code will be available at \url{https://github.com/suweijian1996/DPOFusion}

\end{abstract} 
\vspace{-5mm}
\section{Introduction}
\label{sec:intro}

\begin{figure}[!t]
	\centering
	\includegraphics*[width=1\linewidth]{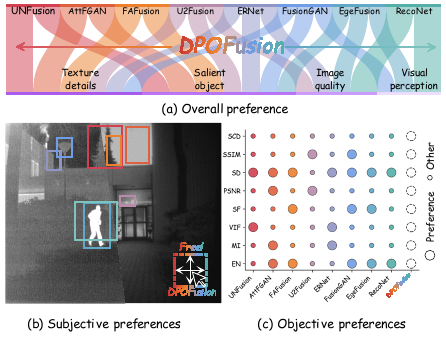}
	\caption{Preference heterogeneity in IVIF. (a) Overall preferences of various methods. (b) Preferred regions of various methods. (c) Metric preferences of various methods.}
        \label{motivation}
        \vspace{-7mm}
\end{figure}

Infrared and visible image fusion (IVIF) seeks to integrate complementary spectral information to enhance human perception and boost performance in various downstream vision tasks~\cite{liu2024infrared, zhang2021deep}. However, different applications and users have heterogeneous requirements for the fused output. As illustrated in Figure~\ref{fig:teaser}, users may prioritize overall image quality, natural appearance, or suitability for downstream tasks such as object detection and segmentation. While learning-based methods have demonstrated preference-aware fusion—capable of aligning with perceptual, regional, or metric-based preferences (Figure~\ref{motivation})—they typically require separate models and complex training strategies for each specific demand, limiting flexibility and scalability. This highlights a critical gap: developing a single, unified fusion framework that can adaptively satisfy diverse user preferences remains an open challenge.

Recently, direct preference optimization (DPO) has emerged as a powerful approach for aligning models, initially demonstrated in large language models (LLMs)~\cite{rafailov2023direct, xiao2024cal, yang2025mitigating}. DPO fine-tunes a policy model directly on a static dataset of preference pairs (\textit{i.e.}, preferred \textit{vs.} rejected outputs), eliminating the need for an explicit reward model and enabling a more stable and efficient alignment process. More recently, DPO has been successfully applied to optimizing diffusion models for image generation, improving both visual fidelity and prompt adherence~\cite{wallace2024diffusion, huang2025patchdpo, jin2025focusdpo}. A key advantage of DPO is that it can adapt a foundational model, such as a latent diffusion model (LDM)~\cite{rombach2022high}, using collected preference feedback, without requiring complex architectural changes or cumbersome joint-optimization pipelines with downstream models~\cite{xu2024dpo, wu2025rethinking}.

Inspired by DPO, we aim to build a unified fusion network capable of adapting to heterogeneous fusion requirements from both human users and downstream machine vision tasks. With only minimal fine-tuning, the original model can be customized to satisfy different fusion requirements, whether driven by user preferences or downstream tasks. We propose DPOFusion, which integrates a property-aligned latent diffusion model (PALDM) and a preference-controlled latent diffusion model (PCLDM). The PALDM is designed to produce diverse, high-quality fusion candidates. We first train a prior latent fusion model to guide the PALDM and introduce a joint conditional denoising loss that aligns generated outputs with property-descriptive text prompts, creating a rich pool of candidates for annotation or task evaluation. The PCLDM further enables fine-grained adjustment of the fusion results according to preference signals. To support this, we introduce instance direct preference optimization (IDPO), which applies preference supervision only within a masked region of interest while enforcing strict pixel-level consistency elsewhere, achieving a precise balance between localized adaptation and global fidelity. Together, the PALDM for candidate generation and the PCLDM trained with IDPO constitute the DPOFusion framework, offering a unified and flexible solution for both human-driven and task-driven fusion demands.

We validate DPOFusion on real-world datasets under a diverse set of heterogeneous demands. As shown in Figure~\ref{fig:teaser}c, it consistently aligns fusion results with the preferences of various evaluators, including human visual perception (RLHF), vision-language model instructions (RLVF), and downstream task models for object detection (RLDF-OD) and segmentation (RLDF-Seg). These results demonstrate that DPOFusion can flexibly adapt to human and machine preferences across different tasks, truly achieving ``fusion in your way'' with different prompts.
\section{Related work}
\label{sec:formatting}
\subsection{Diffusion based multi-modal image fusion}
Diffusion models excel at image generation and are widely used in the field of computer vision~\cite{croitoru2023diffusion, huang2025diffusion, tang2025mask}. Recently, diffusion models have been applied to the field of multi-modal image fusion, including unconditional~\cite{zhao2023ddfm, yi2024diff}, text-guided~\cite{yi2024text, zhang2024text}, and latent space diffusion models~\cite{yue2023dif}.
Unconditional diffusion models incorporate conditional constraints during sampling to guide the preservation of source image features~\cite{zhao2023ddfm,yi2024diff}. Text-guided diffusion models align textual prompts with the source image scene, indicating the direction of fused results~\cite{cao2023ddrf,liang2025fusioninv}. Latent space diffusion models operate directly on latent features extracted from source images, facilitating tasks such as color restoration, feature fusion, and image enhancement~\cite{yue2023dif, shi2024vdmufusion}.

Despite significant progress in multi-modal image fusion, existing models primarily excel in specific tasks, such as enhancing visual quality, addressing downstream tasks, or optimizing certain evaluation metrics.
Nevertheless, existing models are limited to specific fusion tasks and cannot adaptively generate results according to diverse human or machine requirements. Developing a unified framework requires bridging the feature gaps across different task-specific demands, enabling preference-aware fusion to meet heterogeneous objectives and user-defined criteria.

\subsection{Conditional and controllable diffusion models}
Controllable diffusion models synthesize images consistent with diverse user instructions, including textual prompts and spatial masks.
As a representative foundation model, latent diffusion model (LDM)~\cite{rombach2022high} performs latent-space denoising conditioned on multi-modal inputs, providing controllable and high-quality image synthesis.
Building on the effectiveness of stable diffusion, numerous applications have emerged in text-to-image generation~\cite{zhou2023shifted, xue2023raphael, xu2024ufogen}, image editing~\cite{kawar2023imagic, zhang2023sine, mou2024diffeditor}, and inpainting~\cite{lugmayr2022repaint, anciukevivcius2023renderdiffusion}. Concurrently, controllable adaptations have been introduced to enhance LDM's precision~\cite{rombach2022high}. ControlNet~\cite{zhang2023adding}, for instance, injects trainable, zero-convolution-connected blocks into the LDM encode to achieve fine-grained control over the output.
Nevertheless, the scarcity of multi-modal datasets and training strategies have so far hindered multi-modal image fusion models to produce controllable results.

\subsection{Direct preference optimization}

DPO is a fine-tuning framework originally introduced for unsupervised large language models (LLMs)~\cite{rafailov2023direct}. Different from traditional reinforcement learning from human feedback methods that rely on an explicit reward model, DPO formulates preference alignment as a single-stage policy optimization problem, enabling a more stable and computationally efficient fine-tuning process~\cite{zeng2024token, NEURIPS2024_ea888178, xiao2024cal}. This objective directly optimizes the policy to increase the log-probability of preferred (winning) samples while decreasing the log-probability of rejected (losing) samples, relative to a fixed reference policy. The objective function for DPO is defined as
\begin{align}
\footnotesize
\nonumber
  \mathcal{L}_{\text{DPO}}(\theta)=
   &-\mathbb{E}_{c, z_0^w, z_0^l}\\ &\left[
   \log \sigma\left(\beta\log  \frac{p_\theta(z_0^w | c)}{p_{\text{ref}}(z_0^w | c)}-\beta\log  \frac{p_\theta(z_0^l | c)}{p_{\text{ref}}(z_0^l | c)}  \right) \right].
  \label{eq:dpo}
\end{align}

On the basis, diffusion-DPO~\cite{wallace2024diffusion} reformulates the DPO framework to incorporate the likelihood nature of diffusion models, fine-tuning stable diffusion XL-1.0~\cite{podell2023sdxl} to enhance both visual performance and prompt alignment in text-to-image generation. 
Recent adaptations have extended DPO to finer-grained control by shifting the optimization objective from a global, image-level assessment to specific, content-aware local regions or patches~\cite{huang2025patchdpo, jin2025focusdpo}.
These works utilize DPO for fine-tuning text-to-image generation tasks, demonstrate the great potential of DPO in image processing.
Motivated by this, we introduce DPO into the multi-modal image fusion task to achieve preference and task-driven fusion results.
\section{Motivation}


Leveraging DPO in unsupervised infrared–visible image fusion—which aims to align fusion model with human-centric or task-specific objectives—faces two fundamental challenges. First, the absence of ground-truth supervision makes the fusion result highly under-constrained, producing a vast solution space where most candidates fail to achieve high perceptual fidelity and semantic accuracy. Second, preference-adaptive fusion, which requires the model to accommodate distinct perceptual or semantic biases in a localized region, often perturbs shared network parameters, unintentionally affecting unrelated areas. Consequently, the fusion results exhibit inconsistent global structure and degraded visual coherence, underscoring the fusion model to balance local adaptability with global consistency across heterogeneous preference domains.

We observe that even though users may have diverse preferences for fusion results, their criteria for evaluating fusion quality are generally consistent. Motivated by this observation, we propose a preference-guided fine-tuning framework to address the aforementioned challenges. 
Specifically, we first collect a set of high-quality fusion results from a pre-trained model, which serve as a foundation for preference adaptation.
These fusion results are then annotated by users or downstream tasks to explicitly indicate their preferences. Based on the annotated preferences, the model is further fine-tuned to generate fusion outputs that align with user-specific preferences in determined regions, while preserving global consistency across the entire image.

To achieve this, DPOFusion follows a three-stage process. First, PALDM generates multiple high-quality fusion results. Next, a preference data collection module is used for different tasks to explicitly capture region-specific demands. Finally, PCLDM constructs instance-level direct preferences for fine-tuning, enabling adaptive fusion outputs according to localized user or task requirements.

\begin{figure*}[!t]
	\centering
	\includegraphics*[width=1\linewidth]{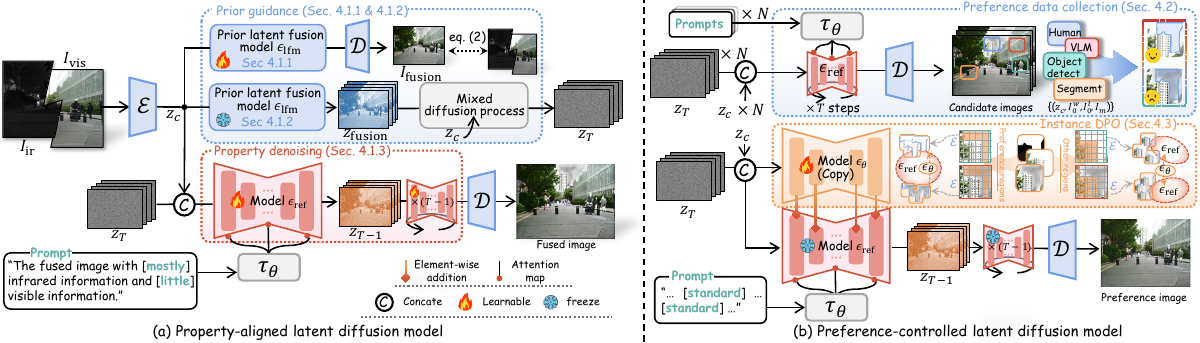}
	\caption{The overall architecture of DPOFusion. It consists of PALDM and PCLDM. PALDM provides candidate fusion samples, and PCLDM fine-tunes the features to generate the preference-aligned fusion result.}
        \label{DPOarch}
        \vspace{-8mm}
\end{figure*}
\section{Methodology}
This section details our DPOFusion architecture, comprising the property-aligned latent diffusion model, the preference-controlled latent diffusion model, and the data collection pipeline, as illustrated in Figure~\ref{DPOarch}.
\subsection{Property-aligned latent diffusion model}
This subsection details the architecture and training process of the property-aligned latent diffusion model and the prior latent fusion model.

\noindent\textbf{Prior latent fusion model.}
To construct supervisory data for PALDM training, we first train a prior latent fusion model, denoted as $\epsilon_\text{lfm}$ (Figure~\ref{DPOarch}a). Specifically, given a pair of source images $I_\text{ir}, I_\text{vis} \in \mathbb{R}^{H \times W \times 3}$, the encoder $\mathcal{E}$ of a pre-trained VAE~\cite{rombach2022high} encodes them into latent representations $z_\text{ir} = \mathcal{E}(I_\text{ir})$ and $z_\text{vis} = \mathcal{E}(I_\text{vis})$. These latents, $z_\text{ir}, z_\text{vis} \in \mathbb{R}^{h \times w \times c}$, are concatenated to form $z_c$. The $\epsilon_\text{lfm}$ network then processes $z_c$ to produce the fused latent $z_\text{fusion} = \epsilon_\text{lfm}(z_c)$. Finally, the VAE decoder $\mathcal{D}$ reconstructs the fused image as $I_\text{fusion} = \mathcal{D}(z_\text{fusion})$. 

The $\epsilon_\text{lfm}$ network employs the Restormer~\cite{zamir2022restormer} architecture, which is optimized via the following fusion loss, \textit{i.e.},
\begin{align}
  \nonumber
\mathcal{L}_{\text{fusion}}=& \sigma_1|\max(I_\text{ir}, I_\text{vis})-I_\text{fusion}| \\&+ \sigma_2|\max(\nabla I_\text{ir}, \nabla I_\text{vis})-\nabla I_\text{fusion}|,  
\end{align}
where $\nabla$ denotes the Sobel operator.

\noindent\textbf{Denoising objective for latent diffusion.} Our PALDM model $\epsilon_{\text{ref}}$ can be interpreted as a sequence of denoising autoencoders trained to predict a clean latent $z_0$ from its noisy version $z_t$ (Figure~\ref{DPOarch}a). 
Note that $\epsilon_{\text{ref}}$ is conditioned on multiple inputs, including the noisy latent $z_t$, the concatenated source features $z_c$, the text prompt $c$, and the timestep $t$.
Specifically, the text prompt $c$ is encoded using a pre-trained CLIP ViT-L/14~\cite{radford2021learning} text encoder $\tau_\theta$ to produce the corresponding text embeddings.
The corresponding denoising objective $\mathcal{L}_\text{denoise}$ is formulated as
\begin{align}
\small
\mathcal{L}_\text{denoise}(z_0, c) = \left\| \epsilon - \epsilon_\text{ref}(z_{t}, z_{c}, c, t) \right\|^2_2,
\label{eq:denoise}
\end{align}
where $\epsilon \sim \mathcal{N}(0,I)$ is the ground-truth noise sampled from a standard normal distribution.

\noindent\textbf{Joint conditional loss.} 
To generate fused samples exhibiting diverse preference data collection, we introduce varied latent targets. 
With the prior $\epsilon_\text{lfm}$ as a base, we produce latent features $z_\text{fusion}^{\prime}$ by first defining $N$ discrete levels and sampling a random level $k \sim \mathcal{U}\{0, \dots, N-1\}$. 
Subsequently, the interpolation coefficient $\alpha = k / (N-1)$ governs a linear interpolation between the source latents $z_\text{ir}$, $z_\text{vis}$, and the prior fused latent $z_\text{fusion}$, denoted as
\begin{align}
\small
z_\text{fusion}^{\prime} = \frac{1}{2} \left( \alpha \cdot z_\text{ir} + (1 - \alpha) \cdot z_\text{vis} \right) + \frac{1}{2} z_\text{fusion}.
\label{eq:z_prime}
\end{align}

This joint loss is particularly well-suited for property alignment, as it trains $\epsilon_{\text{ref}}$ to (i) denoise the standard latent $z_\text{fusion}$ conditioned on the general text prompt $c_t$, and (ii) simultaneously denoise the property-interpolated latent $z_\text{fusion}^{\prime}$ conditioned on its specific property prompt $c_t^{\prime}$.
 Based on the above analysis, the final loss is defined as
\begin{align}
\label{lc}
\small
\mathcal{L}_c = \mathbb{E}_{z_c, k, t, \epsilon} \left[ \mathcal{L}_\text{denoise}(z_\text{fusion}, c_{t}) 
+ \lambda \mathcal{L}_\text{denoise}(z_\text{fusion}^{\prime}, c_{t}^{\prime}) \right],
\end{align}
where $\lambda$ is a balancing coefficient.

\subsection{Preference data collection}
\label{sec:dataset_collection}
Our PCLDM is trained on a heterogeneous preference dataset $D=\{(z_{c},I_{0}^{w},I_{0}^{l},I_{m})\}$. As illustrated in Figure~\ref{DPOarch}b, we collect this data from three distinct sources, including human annotators, task-specific models, and advanced VLMs.
Further details on the annotation process are provided in the Appendix.

\noindent\textbf{Region-specific preference collection} 
is designed for fine-grained, localized alignment. It provides a specific preference region mask $I_m$ along with the preferred $I_0^w$ and rejected $I_0^l$ samples. This mask allows our IDPO loss to apply preference optimization exclusively to the target area while preserving image quality elsewhere. We utilize this function to collect data for subjective human preferences and task-driven segmentation.

\noindent\textbf{Global preference collection} is used for tasks where collecting precise instance masks is difficult. For these tasks, we design a sample filtering pipeline to identify preferred $I_0^w$ and rejected $I_0^l$ samples, and simultaneously extract their corresponding local preference patches. The mask $I_m$ is set to cover the entire patch. This function is used to collect data from advanced VLMs and for the object detection task.
\begin{figure*}[!t]
	\centering
	\includegraphics*[width=1\linewidth]{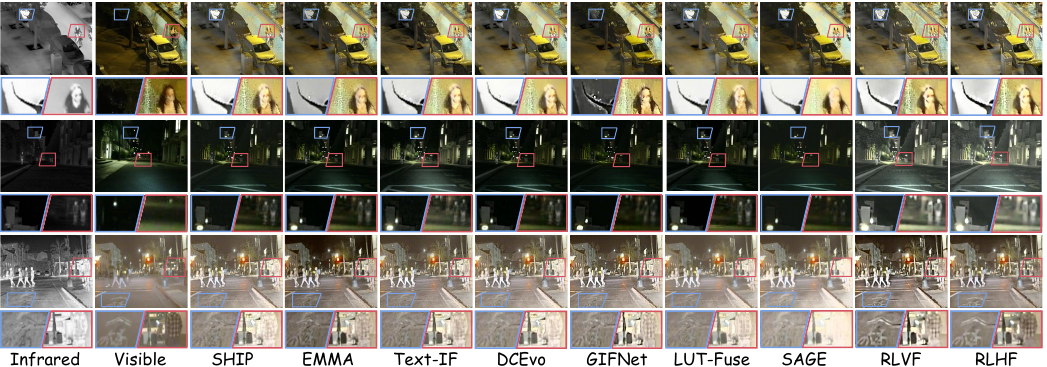}
	\caption{Qualitative comparison of DPOFusion (fine-tuned with human and VLM feedback) against state-of-the-art fusion methods on the LLVIP, MSRS, and RoadScene datasets.}
        \label{visual}
        \vspace{-2mm}
\end{figure*}

\begin{table*}[htbp]
\footnotesize
  \centering
  \caption{Quantitative comparison of our methods against state-of-the-art fusion methods on the LLVIP, MSRS, and RoadScene datasets. The best values are in \textbf{bold}, and the second-best values are \underline{underlined}.}
  \renewcommand{\arraystretch}{1.1} 
\setlength{\tabcolsep}{1.1mm}{
       \begin{tabular}{ccccccccccccccccc}
    \hline
    \multicolumn{1}{c}{\multirow{2}[1]{*}{Methods}} & \multicolumn{1}{c}{\multirow{2}[1]{*}{Reference}} & \multicolumn{5}{c}{\textbf{LLVIP Dataset}}    & \multicolumn{5}{c}{\textbf{MSRS Dataset}}     & \multicolumn{5}{c}{\textbf{RoadScene Dataset}} \\
    \cmidrule(lr){3-7} \cmidrule(lr){8-12} \cmidrule(lr){13-17} 
          &       & EN & SD & AG & MUS & CNN & EN & SD & AG & MUS & CNN & EN & SD & AG & MUS & CNN \\
          \hline
    U2Fusion & TPAMI$^{22}$ & 7.441  & 50.961  & 4.473  & 56.384  & 0.649  & 5.042  & 20.342  & 2.264  & 32.795  & 0.379  & 6.998  & 35.743  & 5.125  & 42.170  & 0.444  \\
    DDFM  & ICCV$^{23}$ & 7.001  & 38.018  & 4.351  & 40.997  & 0.619  & 6.125  & 28.479  & 2.955  & 29.457  & 0.395  & 7.225  & 41.676  & 5.098  & 33.274  & 0.469  \\
    SHIP  & CVPR$^{24}$ & 7.386  & 49.104  & 5.044  & 56.331  & 0.648  & 6.429  & 41.152  & 3.936  & 36.570  & 0.442  & 7.148  & 42.969  & 5.963  & \cellcolor{Best}\textbf{46.976} & \cellcolor{Secondbest}\underline{0.556} \\
    EMMA  & CVPR$^{24}$ & 7.429  & 50.199  & 4.564  & 56.683  & 0.640  & 6.723  & 44.591  & 3.788  & 35.977  & 0.374  & \cellcolor{Secondbest}\underline{7.499} & \cellcolor{Best}\textbf{56.300} & 6.163  & 39.655  & 0.425  \\
    Text-IF & CVPR$^{24}$ & 6.726  & 36.521  & 3.619  & 49.664  & 0.555  & 7.040  & 44.289  & 3.784  & 35.843  & 0.398  & 7.273  & 45.749  & 6.333  & 45.557  & 0.530  \\
    DCEvo & CVPR$^{25}$ & 6.938  & 44.423  & 5.345  & 51.853  & 0.636  & 6.626  & 42.942  & 3.728  & 36.602  & 0.439  & 7.167  & 44.128  & 5.318  & 43.498  & 0.538  \\
    GIFNet & CVPR$^{25}$ & 7.355  & 49.478  & 3.901  & 55.277  & 0.644  & 5.944  & 32.971  & 3.375  & 38.902  & \cellcolor{Secondbest}\underline{0.538} & 7.342  & 48.020  & \cellcolor{Secondbest}\underline{6.413} & 45.511  & \cellcolor{Best}\textbf{0.605}\\
    LUT-Fuse & ICCV$^{25}$ & 6.894  & 51.270  & 3.960  & 49.352  & 0.630  & 6.671  & 42.781  & 3.770  & 38.200  & 0.424  & 6.959  & 42.569  & 4.454  & \cellcolor{Secondbest}\underline{45.820}& 0.475  \\
    SAGE & CVPR$^{25}$ & 7.355  & 50.836  & 4.550  & 53.370  & 0.634  & 6.004  & 36.338  & 3.168  & 37.418  & 0.398  & 7.007  & 42.213  & 3.888  & 37.124  & 0.435  \\
    \cdashline{1-17}
    \rowcolor{gray!10}
    \textbf{RLVF} & Ours  & \cellcolor{Secondbest}\underline{7.554} & \cellcolor{Secondbest}\underline{53.299} & \cellcolor{Secondbest}\underline{5.439} & \cellcolor{Secondbest}\underline{57.280} & \cellcolor{Best}\textbf{0.660} & \cellcolor{Best}\textbf{7.203} & \cellcolor{Best}\textbf{56.614} & \cellcolor{Best}\textbf{5.782} & \cellcolor{Best}\textbf{39.684} & 0.524  & \cellcolor{Best}\textbf{7.574} & \cellcolor{Secondbest}\underline{53.323} & \cellcolor{Best}\textbf{8.622} & 41.431  & 0.472  \\   
 \rowcolor{gray!10}
    \textbf{RLHF} & Ours  & \cellcolor{Best}\textbf{7.725} & \cellcolor{Best}\textbf{61.911} & \cellcolor{Best}\textbf{5.977} & \cellcolor{Best}\textbf{57.295} & \cellcolor{Secondbest}\underline{0.655} & \cellcolor{Secondbest}\underline{7.138} & \cellcolor{Secondbest}\underline{53.385} & \cellcolor{Secondbest}\underline{5.601} & \cellcolor{Secondbest}\underline{39.140} & \cellcolor{Best}\textbf{0.545} & 7.210  & 40.535  & 5.957  & 43.785  & 0.539  \\
    \hline
    \end{tabular}}
  \label{tab:three}%
  \vspace{-5mm}
\end{table*}%

\subsection{Preference-controlled latent diffusion models}
This subsection details the architecture of the preference-controlled latent diffusion model and its training using instance direct preference optimization.

\noindent\textbf{Controllable diffusion models.}
The PCLDM $\epsilon_\theta$ is a finetuning model to steer the standard latent features $z_\text{fusion}$ addressing various user preferences. 
During the PCLDM training stage, the PALDM $\epsilon_\text{ref}$, is frozen to provide the foundational fusion capabilities for the framework~(Figure~\ref{DPOarch}b). We denote the frozen PAMDL model as $\mathcal{F}_\text{ref}(\cdot; \Theta_\text{ref})$, with frozen parameters $\Theta_\text{ref}$.
The PCLDM $\epsilon_\theta$ is initialized from a duplicated model $\mathcal{F}_{\theta}(\cdot; \Theta_{\theta})$, 
which inherits the fusion priors of $\epsilon_{\text{ref}}$, enabling rapid preference-specific fine-tuning.

The frozen model $\mathcal{F}_{\text{ref}}$ is conditioned on the input feature $x$ and a general text prompt $c_t$, while the trainable counterpart $\mathcal{F}_{\theta}$ takes the same $x$ but a preference-aligned prompt $c_t^{\prime}$. The trainable $\mathcal{F}_{\theta}$ is connected to the frozen reference $\mathcal{F}_{\text{ref}}$ through a zero-initialized $1{\times}1$ convolution layer $\mathcal{Z}(\cdot; \Theta_z)$. The output $y_p$ of the combined module is obtained by summing the outputs of the two branches, \textit{i.e.},
\begin{align}
   y_{p} = \mathcal{F}_\text{ref}(x, c_{t}; \Theta_\text{ref}) + \mathcal{Z}(\mathcal{F}_{\theta}(x, c_{t}^{\prime}; \Theta_{\theta}); \Theta_z), 
\end{align}

\begin{table*}[!t]
\footnotesize
  \centering
  \caption{Quantitative comparison of our methods (RLDF-Seg for MSRS, RLDF-OD for M3FD) with state-of-the-art fusion methods for object segmentation and object detection.}
      \renewcommand{\arraystretch}{1.1} 
\setlength{\tabcolsep}{0.9mm}{
    \begin{tabular}{c|ccccccccc|ccccccccc}
       \hline
    Dataset & \multicolumn{9}{c|}{\textbf{MSRS Dataset}}                                              & \multicolumn{9}{c}{\textbf{M3FD Dataset}} \\
    \hline
    Methods & Car   & Person & Bike  & Curve & Car-S. & Guard. & Cone  & Bump  & mIoU  & Bus &Car & Lamp & Moto. & People & Truck & @.5:.95 & @.5 & @.75 \\
    \hline
        {U2Fusion} & \cellcolor{Best}\textbf{80.35} & \cellcolor{Secondbest}\underline{63.15} & 60.18 & 23.75 & \cellcolor{Secondbest}\underline{43.16} & 38.28 & 25.20  & \cellcolor{Best}\textbf{61.03} & 54.71 & 62.48  & \cellcolor{Best}\textbf{60.79} & \cellcolor{Best}\textbf{25.77} & 20.40  & \cellcolor{Secondbest}\underline{43.02} & 44.59  & 42.84  & 66.24  & 43.62  \\
    {DDFM} & 78.39 & 60.48 & 56.18 & 22.05 & 43.02 & 34.46 & 23.46 & \cellcolor{Secondbest}\underline{56.93} & 52.45 & 63.75  & 60.04  & 23.74  & 20.90  & 42.03  & 41.67  & 42.02  & 64.38  & 43.90  \\
    {SHIP} & 78.07 & 60.25 & 57.66 & 26.33 & 37.61 & 46.12 & 24.7  & 53.12 & 53.44 & 62.43  & 58.24  & \cellcolor{Secondbest}\underline{25.46} & \cellcolor{Secondbest}\underline{23.97} & 40.40  & 41.35  & 41.98  & 65.54  & \cellcolor{Secondbest}\underline{44.18} \\
    {EMMA} & 80.06 & 60.84 & \cellcolor{Best}\textbf{60.53} & 30.36 & 40.88 & 48.74 & 29.01 & 50.03 & 55.30 & 60.07  & 57.35  & 21.82  & 17.89  & 38.38  & 40.81  & 39.39  & 63.24  & 39.92  \\
    {Text-IF} & 79.71 & 59.18 & 58.23 & 21.85 & 41.42 & 46.72 & 24.38 & 52.48 & 53.45 & \cellcolor{Best}\textbf{64.50} & 57.86  & 22.05  & 22.52  & 39.85  & 44.50  & 41.88  & 66.28  & 42.81  \\
    {DCEvo} & \cellcolor{Secondbest}\underline{80.33} & \cellcolor{Best}\textbf{63.56} & \cellcolor{Secondbest}\underline{60.52} & 28.91 & 41.96 & 47.82 & 27.79 & 52.72 & \cellcolor{Secondbest}\underline{55.66} & 62.00  & \cellcolor{Secondbest}\underline{60.78} & 23.02  & 22.53  & \cellcolor{Best}\textbf{43.57} & 45.42  & \cellcolor{Secondbest}\underline{42.89} & \cellcolor{Best}\textbf{68.44} & 42.93  \\
    {GIFNet} & 79.90  & 59.43 & 59.59 & 23.15 & \cellcolor{Best}\textbf{44.58} & 41.00    & 25.62 & 52.80  & 53.70  & 61.44  & 58.59  & 24.25  & 21.35  & 40.91  & 39.04  & 40.93  & 63.76  & 41.94  \\
    {LUT-Fuse} & 79.64 & 62.94 & 59.66 & \cellcolor{Secondbest}\underline{30.88} & 40.13 & \cellcolor{Secondbest}\underline{48.93} & \cellcolor{Secondbest}\underline{29.29} & 42.89 & 54.62 & 61.62  & 58.28  & 19.61  & 21.95  & 41.61  & \cellcolor{Secondbest}\underline{46.85} & 41.65  & 63.08  & 43.49  \\
    {SAGE} & 78.08 & 54.86 & 57.46 & 19.51 & 37.67 & 42.34 & 19.92 & 53.09 & 51.10  & 61.81  & 59.02  & 23.27  & 18.22  & 41.09  & 36.95  & 40.06  & 62.93  & 42.75  \\
        \cdashline{1-17}
    \rowcolor{gray!10}
    \textbf{RLDF$^*$} & 80.13 & 62.08 & 59.05 & \cellcolor{Best}\textbf{31.79} & 41.19 & \cellcolor{Best}\textbf{49.40} & \cellcolor{Best}\textbf{33.09} & 49.73 & \cellcolor{Best}\textbf{55.96} & \cellcolor{Secondbest}\underline{64.07} & 58.85  & 23.26  & \cellcolor{Best}\textbf{26.27} & 39.45  & \cellcolor{Best}\textbf{48.50} & \cellcolor{Best}\textbf{43.40} & \cellcolor{Secondbest}\underline{66.48} & \cellcolor{Best}\textbf{46.03} \\
    \hline
    \end{tabular}}
        \vspace{-3mm}
  \label{tab:down}%
\end{table*}%
It is noted that zero initialization protects the backbone's foundational fusion performance from harmful noise during the initial fine-tuning by eliminating random noise as gradients in the initial training steps.

\noindent\textbf{Instance direct preference optimization.} As mentioned earlier, to align localized regions with user preferences, the training dataset is defined as $D = \{(z_c, I_0^w, I_0^l, I_m)\}$, where each sample consists of the source latent $z_c$, a preference pair $(I_0^w, I_0^l)$ labeled by evaluators with $I_0^w \succ I_0^l$,  and a corresponding preference mask $I_m$ indicating the region of interest.

We propose IDPO to achieve precise local preference alignment while preserving global visual cues. The IDPO loss $\mathcal{L}_{\text{IDPO}}$ is formulated as a weighted objective that (i) applies a preference alignment loss exclusively within the masked region $I_m$, and (ii) enforces a strict pixel-level consistency loss outside the mask, constraining those regions to remain consistent with the output of the frozen $\epsilon_{\text{ref}}$.

We define $\mathcal{P}_w$ and $\mathcal{P}_l$ as the difference between the squared error of $\epsilon_\theta$ and $\epsilon_\text{ref}$, computed within the masked region for the preferred and rejected samples, \textit{i.e.},
\begin{equation}
\begin{cases}
    \mathcal{P}_w = \| (\epsilon^w - \epsilon_\theta^w) \odot z_m \|_2^2 - \| (\epsilon^w - \epsilon_{\text{ref}}^w) \odot z_m \|_2^2 \\
    \mathcal{P}_l = \| (\epsilon^l - \epsilon_\theta^l) \odot z_m \|_2^2 - \| (\epsilon^l - \epsilon_{\text{ref}}^l) \odot z_m \|_2^2
\end{cases}
,
\label{eq:p_terms}
\end{equation}
where $\odot$ denotes element-wise product. $\epsilon^w$ and $\epsilon^l$ are the ground-truth noise vectors sampled from $\mathcal{N}(0,I)$. $\epsilon^*_\text{ref}$ and $\epsilon^*_\theta$ denote $\epsilon^*_\text{ref}(z_t^*, z_c, c_t, t)$ and $\epsilon^*_\theta(z_t^*, z_c, c_t^{\prime}, t)$, respectively. $z_m$ is the latent mask obtained by nearest-neighbor downsampling of $I_m$.

The $\mathcal{O}_w$ and $\mathcal{O}_l$ represent the pixel-level constraint imposed by $\epsilon_\text{ref}$ on the other regions.
\begin{equation}
\begin{cases}
    \mathcal{O}_w = \| (\epsilon_\theta^w - \epsilon_{\text{ref}}^w) \odot z_m^c \|_2^2 \\
    \mathcal{O}_l = \| (\epsilon_\theta^l - \epsilon_{\text{ref}}^l) \odot z_m^c \|_2^2
\end{cases}
,
\label{eq:n_terms}
\end{equation}
where $z_m^c$ is the latent mask for the other regions.

Based on the above global and local fusion constraints, the model is jointly trained with multiple loss terms. The $\mathcal{L}_\text{IDPO}$ is calculated as
\begin{align}
\mathcal{L}_\text{IDPO} &= \mathbb{E}_{\substack{
(z_c, z_0^w, z_0^l, z_m) \sim \mathcal{D}, t \sim \mathcal{U}\{1,\dots,T\} 
, \epsilon^w, \epsilon^l \sim \mathcal{N}(0, {I})
}}\nonumber \\&
\left[
    \underbrace{
        -\log \sigma \left( -\beta_t \cdot (\mathcal{P}_w - \mathcal{P}_l) \right)
    }_{\text{Preference region}}
    + \mu \cdot
    \underbrace{
        \left( \mathcal{O}_w + \mathcal{O}_l \right)
    }_{\text{Other region}}
\right],
\label{eq:idpo_final_annotated}
\end{align}
where {hyperparameter $\beta_t$} is the divergence penalty parameter, which scales the implicit reward and controls the regularization strength, following the DPO framework. $\mu$ is a balancing coefficient to weigh the preference-alignment loss against the consistency loss in the other regions. Following $T$ preference-controlled denoising steps, the VAE decoder $\mathcal{D}$ reconstructs the final preference-aligned image $I_\text{pref} = \mathcal{D}(z_0)$.

\begin{figure*}[!t]
	\centering
	\includegraphics*[width=1\linewidth]{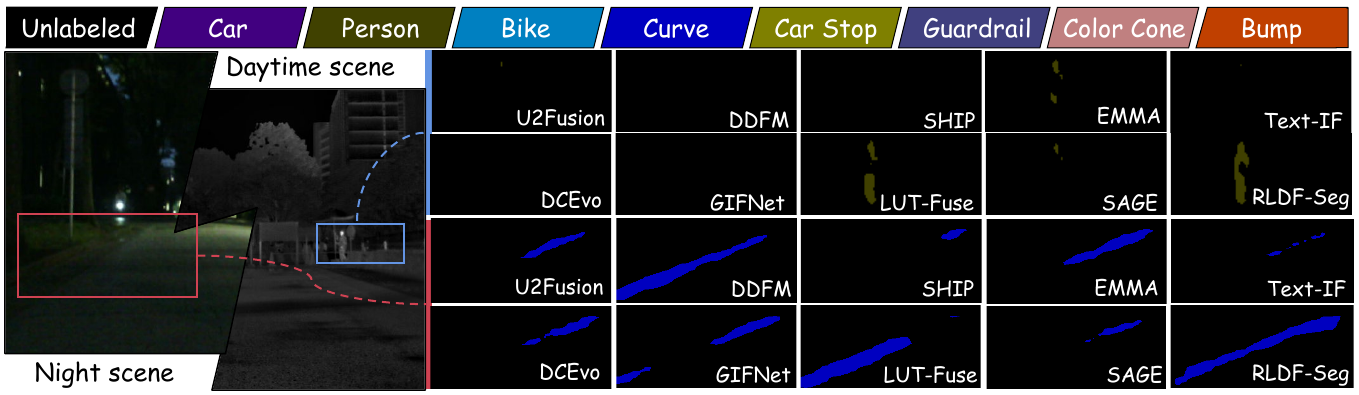}
	\caption{Qualitative comparison of semantic segmentation performance between our RLDF-Seg and state-of-the-art methods on the MSRS dataset.}
        \label{Seg}
        \vspace{-5mm}
\end{figure*}

\begin{figure*}[!t]
	\centering
	\includegraphics*[width=1\linewidth]{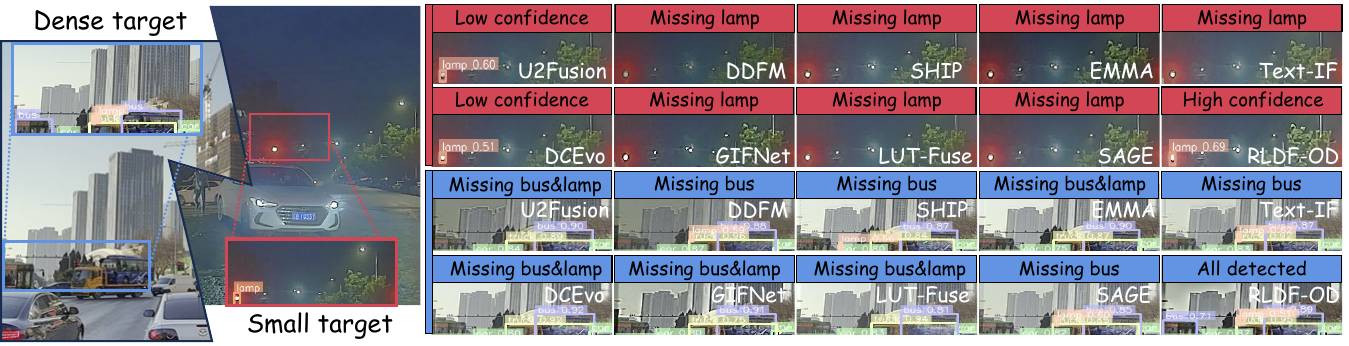}
	\caption{Qualitative comparison of object detection performance between our RLDF-OD and state-of-the-art methods on the M3FD dataset.}
        \label{OD}
        \vspace{-8mm}
\end{figure*}
\section{Experiments}
\subsection{Setting up}

\noindent \textbf{Heterogeneous demands indication.}
To evaluate the fusion results in heterogeneous demands, we provide four types of prompts: (1) local-region visual preference of human (noted as RLHF), (2) preference of the vision language model (noted as RLVF), (3) detection results (noted as RLDF-OD), and (4) segmentation results (noted as RLDF-Seg).  The first type evaluates DPOFusion can align with human local visual perception, the second examines the model's adaptability to understand the vision language model, and the last two assess its applicability to machine-driven downstream tasks.
Specifically, the RLVF task uses the output of QWEN3-Omni-Think~\cite{xu2025qwen3} as the prompt, the RLDF-OD and RLDF-Seg evaluate the response to yolov11~\cite{khanam2024yolov11} and SegFormer~\cite{xie2021segformer}, respectively.

\noindent\textbf{Hyper-parameter and training details.} As mentioned earlier, the PALDM is employed to generate candidate fusion results that represent potential intended outcomes. Subsequently, the PCLDM optimizes the fusion results based on selected preference image pairs. 
The unified PALDM model is pre-trained using checkpoint with the lowest validation loss. During PCLDM training, we set the learning rate to $1\times10^{-5}$ and use a batch size of $8$. For each specific prompt, the PCLDM is fine-tuned from this base model for $20$ epochs.

{We set $ \beta_t = 10$ for RLHF, RLVF, and RLDF-Seg, $\beta_t = 500$ for RLDF-OD task.} 
The $\lambda $ is set to 2. 
For the $\mathcal{L}_\text{fusion}$, the $\sigma_1$ and $\sigma_2$ are set to 4 and 10, respectively. 
All models are trained on 2 $\times$ GeForce RTX 4090 GPUs, with inference performed on a single GPU. Other hyperparameters in diffusion model are set same as research~\cite{rombach2022high, zhang2023adding}.

\noindent\textbf{Datasets.}
We use the LLVIP~\cite{jia2021llvip} dataset for training, with images cropped into $256 \times 256$ patches.  
For PCLDM fine-tuning, the RLHF and RLVF preference datasets are generated from a source pool of 100 images sampled from each of the LLVIP, MSRS~\cite{tang2022piafusion}, and RoadScene~\cite{xu2020fusiondn} datasets. The RLDF training set is composed of 540 images from M3FD~\cite{liu2022target} and 1083 images from MSRS. The test set is a composite benchmark comprising 100 images from LLVIP, 361 images from MSRS, 121 images from RoadScene, and 300 images from M3FD.

\noindent\textbf{Metric.} 
We adopt various evaluation metrics include entropy~(EN), standard deviation (SD), average gradient (AG), MUSIQ~(MUS)~\cite{ke2021musiq}, and CNNIQA~(CNN)~\cite{kang2014convolutional}, where a higher score signifies better fusion quality for all indicators.

\noindent\textbf{SOTA competitors.} 
We compare DPOFusion against nine other state-of-the-art methods, including U2Fusion~\cite{xu2020u2fusion}, DDFM~\cite{zhao2023ddfm}, SHIP~\cite{zhao2024equivariant}, EMMA~\cite{zhao2024equivariant}, Text-IF~\cite{yi2024text}, DCEvo~\cite{liu2025dcevo}, GIFNet~\cite{cheng2025one}, LUT-Fuse~\cite{yi2025lut}, and SAGE~\cite{wu2025every}. {Additional experimental comparisons are provided in the Supplementary materials.}

\subsection{Preference alignment evaluation on IVIF}
This section evaluates quantitative and qualitative results with human perception and language instruction as prompt.

\noindent\textbf{Qualitative comparisons.} 
Qualitative results on three datasets are presented in Figure~\ref{visual}. The results of DPOFusion, after being fine-tuned with VLM and human preference data (termed RLVF and RLHF, respectively), demonstrate superior image quality and align more closely with human visual perception. In the low-light scenario (group 2), only the DPOFusion-based methods successfully reveal the contrast details of small infrared objects. Furthermore, in the complex illumination scenarios of the RoadScene dataset, DPOFusion results (RLVF and RLHF) demonstrate enhanced preservation of texture information.  
Specifically, the RLHF result faithfully retains the complete plaid patterns on clothing, indicating the DPOFusion's capability to capture fine-grained local visual features.

\noindent\textbf{Quantitative comparisons.} 
As shown in Table~\ref{tab:three}, the quantitative results demonstrate that DPOFusion methods achieve state-of-the-art performance. RLVF achieves optimal values on EN, SD, and AG in most cases, indicating its commitment to enhancing texture details. Similarly, RLHF achieves optimal scores for MUSIQ and CNNIQA in most case, suggesting that it aligns closely with human visual perception of overall image. RLHF achieves comparatively lower performance metrics on the RoadScene dataset. This behavior can be attributed to the tendency of human evaluators, in complex scenarios, to prioritize holistic visual coherence over the precise details of localized regions.


\begin{figure}[!t]
	\centering
	\includegraphics*[width=0.9\linewidth]{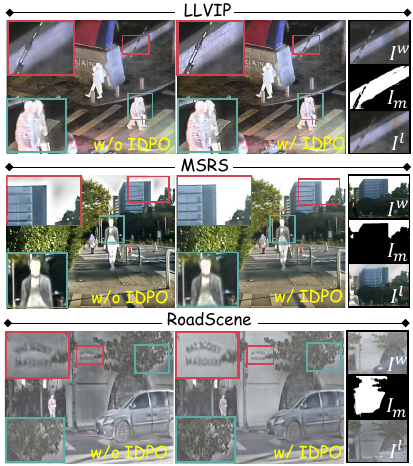}
	\caption{Evaluation of IDPO fine-tuning with RLHF preference alignment.}
 \vspace{-3mm}
\label{pref}
\end{figure}

\begin{table}[!t]
\small
  \centering
  \caption{Quantitative comparison of with different loss components. The best values are shown in \textbf{bold}.}
  \renewcommand{\arraystretch}{1.1} 
    \begin{tabular}{l|ccccc}
    \hline
    Method & {EN} & {SD} & {AG} & {MUS} & {CNN} \\
    \hline
     \rowcolor{CVPR!8}
    \multicolumn{6}{l}{\textcolor{CVPR}{\textit{PALDM denoise loss variants}}} \\ 
    w/ $\mathcal{L}_\text{ST}$ & 7.647  & 58.215  & 5.163  & 55.515  & 0.649  \\
    w/ $\mathcal{L}_\text{ET}$ & 7.470  & 53.612  & 3.782  & 53.848  & 0.608  \\
                  \cdashline{1-6}
        \rowcolor{gray!10}
    \textbf{w/ $\mathcal{L}_c$} & \textbf{7.680}  & \textbf{60.806}  &\textbf{ 5.343 } & \textbf{56.154}  & \textbf{0.659}  \\
    \hline
     \rowcolor{CVPR!8}
    \multicolumn{6}{l}{\textcolor{CVPR}{\textit{PCLDM preference loss variants}}} \\ 
    w/ $\mathcal{L}_\text{DPO}$ &   7.600    & 57.089      &  \textbf{14.215}     &   53.850    &  0.652\\
    w/ $\mathcal{L}_\text{contrast}$ &  7.352     &     51.077  &  8.232     &   55.303    & \textbf{0.694} \\
 
                  \cdashline{1-6}
    \rowcolor{gray!10}
   \textbf{w/ $\mathcal{L}_\text{IDPO}$} & \textbf{7.725} & \textbf{61.911} & 5.977 & \textbf{57.295} & 0.655 \\
    \hline
    \hline
    Method & &\multicolumn{2}{l}{mIoU} & \multicolumn{2}{l}{mAcc}  \\
      \hline
      \rowcolor{CVPR!8}
          \multicolumn{6}{l}{\textcolor{CVPR}{\textit{Segmentation task}}} \\ 
    w/o $\mathcal{L}_\text{IDPO}$ & &\multicolumn{2}{l}{55.21} & \multicolumn{2}{l}{63.52}  \\
    w/ $\mathcal{L}_\text{DPO}$ & &\multicolumn{2}{l}{55.20} & \multicolumn{2}{l}{62.52}  \\
                  \cdashline{1-6}
    \rowcolor{gray!10}
    \textbf{w/ $\mathcal{L}_\text{IDPO}$} & &\multicolumn{2}{l}{\textbf{55.96}} & \multicolumn{2}{l}{\textbf{64.26}}  \\
      \hline    
    \end{tabular}%
    \vspace{-3mm}
  \label{tab:abstudy}%
\end{table}%
\subsection{Downstream task preference alignment}
\noindent \textbf{Semantic segmentation.} The quantitative results for semantic segmentation are presented in the left part of Table~\ref{tab:down}.  
The RLDF-Seg finetunes preference data under the segmentation task, achieves a 0.5\% improvement in \text{mIoU}. Also, Figure~\ref{Seg} shows qualitative visualization results. Notably, instance-level preference finetuning in RLDF-Seg enables effective segmentation of small targets, demonstrating the model's ability to capture fine-grained structural details.

\noindent\textbf{Detection.} The quantitative results of objective detection are shown in the right of Table~\ref{tab:down}. DPOFusion achieves a 4.2\% improvement in mAP, demonstrating the effectiveness of fine-tuning when the detection task is used as the guiding prompt. Figure~\ref{OD} shows the visualization results for the object detection task, where RLDF-OD achieves the best detection performance on dense targets. The results demonstrate that DPOFusion effectively captures the key features that influence the performance of the detection model.

\subsection{Ablation study}
\noindent\textbf{Evaluation of IDPO.} We conduct ablation studies to assess the effectiveness of IDPO. As shown in Figure~\ref{pref}, RLHF with IDPO fine-tuning leads to clear improvements in the fusion results: the model adjusts the user-preferred regions (red boxes) while maintaining high fusion quality in the remaining areas (green boxes). This demonstrates that IDPO enables targeted preference alignment without degrading global fusion performance.

\begin{figure}[!t]
	\centering
	\includegraphics*[width=1\linewidth]{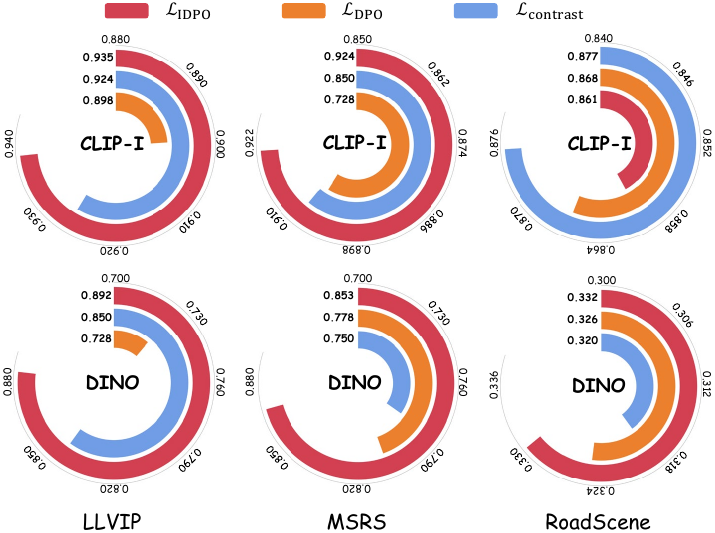}
	\caption{Quantitative comparison of alignment performance for various preference alignment losses.}
            \vspace{-3mm}
        \label{sim}
\end{figure}

\noindent\textbf{Evaluation of loss components.} Table~\ref{tab:abstudy} summarizes the ablation studies for our proposed joint conditional loss and IDPO. We first train PALDM using a single-objective loss $\mathcal{L}_\text{ST}$ (the first term of Eq.~(\ref{lc})) and a multi-objective loss $\mathcal{L}_\text{ET}$ (combining the first and second terms of Eq.~(\ref{lc})). The results demonstrate that the joint conditional loss consistently produces higher-quality candidate fusion images.
We then evaluate PCLDM using DPO ($\mathcal{L}_\text{DPO}$) and contrastive loss ($\mathcal{L}_\text{contrast}$) as alternatives to IDPO. Experimental results show that, across both general fusion and downstream segmentation tasks, IDPO achieves better alignment with user preferences and generates superior fusion results. Figure~\ref{sim} further illustrates the CLIP-I and DINO similarity between the outputs of models trained with different preference losses and the ground-truth preferred images, confirming that IDPO enables more precise preference learning.


\section{Conclusion}
We present DPOFusion, a direct preference optimization framework for infrared and visible image fusion (IVIF) designed to align fused images with heterogeneous demands. By leveraging a property-aligned latent diffusion model (PALDM), our method first generates a diverse, high-quality candidate pool where a joint conditional loss aligns fusion properties with specific text prompts. The proposed instance direct preference optimization (IDPO) strategy effectively handles fine-grained, localized demands by applying preference loss exclusively to masked regions, while a strict consistency loss preserves global image quality.
Extensive experiments demonstrate DPOFusion achieves state-of-the-art fusion quality and precise alignment with diverse evaluators (RLHF, RLVF, RLDF-OD, and RLDF-Seg), verifying DPOFusion's potential as a unified, adaptive solution for real-world IVIF applications.
\section*{Acknowledgments}
This work was supported by the National Key R\&D Program of China under grant No.~2024YFA1012700, the National Natural Science Foundation of China 62502069, 12561015, the Natural Science Foundation of Liaoning Province 2025-MS-007, the Natural Science Foundation of Ningxia Province 2025AAC020002, and the Liaoning Young Elite Scientists Sponsorship Program.
{
    \small
    \bibliographystyle{ieeenat_fullname}
    \bibliography{main}
}

\clearpage
\setcounter{page}{1}
\maketitlesupplementary

\appendix
\section{Methodology details}
This section details the network architecture, hyperparameters, and the preference data annotation process.
\subsection{Network architecture}
We use a Transformer-based U-Net architecture as the backbone of our latent diffusion model. To incorporate semantic guidance, the input text prompts $c_t$ are encoded into semantic embeddings with a pre-trained CLIP ViT-L/14~\cite{radford2021learning, clip} text encoder $\tau_\theta$, denoted as

\begin{align}
    c_\text{emb} = \tau_\theta(c_t).
\end{align}

These embeddings are subsequently injected into the U-Net via cross-attention. Specifically, the intermediate spatial features $\phi(z_t)$ from the U-Net and the text embeddings $c_\text{emb}$ are projected into query $Q$, key $K$, and value $V$ matrices, which is denoted as

\begin{align}
    Q = W_Q \cdot \phi(z_t), \quad K = W_K \cdot c_\text{emb}, \quad V = W_V \cdot c_\text{emb},
\end{align}
where $W_Q$, $W_K$, and $W_V$ are learnable projection matrices. The cross-attention output then guides the denoising process, \textit{i.e.},

\begin{align}
    \text{Attention}(Q, K, V) = \text{Softmax}\left(\frac{QK^T}{\sqrt{d}}\right)V,
\end{align}
where $d$ represents the channel dimension of the keys and queries. This mechanism ensures the fusion output aligns with the specific property. The detailed parameters of PALDM and PCLDM are presented in Table~\ref{supp:arch}.

\begin{table}[htbp]
\small
  \centering
  \caption{Hyperparameters for the LDMs.}
          \renewcommand{\arraystretch}{1.1} 
\setlength{\tabcolsep}{2.5mm}{
    \begin{tabular}{lcc}
    \toprule
      Configuration  & PALDM  & PCLDM  \\
    \midrule
    
    Downsampling     & 4     & 4 \\
    Latent shape ($z$) & $64\times64\times9$ & $64\times64\times9$ \\
    Diffusion steps & 1000  & 1000 \\
    Noise schedule & linear & linear \\
    $N_\text{params}$ & 420.26M & 179.51M \\
    Channels & 224   & 224 \\
    Depth & 2     & 2 \\
    Channel multiplier & 1,2,3,4 & 1,2,3,4 \\
    Attention resolutions & 8,4,2 & 8,4,2 \\
    Head channels & 32    & 32 \\
    Batch size & 8     & 8 \\
    Epochs & 56   & 20\\
    Learning rate & 5e-6 & 1e-5 \\
    \bottomrule
    \end{tabular}}
  \label{supp:arch}%
\end{table}%
\subsection{Preference data collection and annotation}

\label{app:data_coll}
\begin{figure*}[!t]
	\centering
	\includegraphics*[width=1\linewidth]{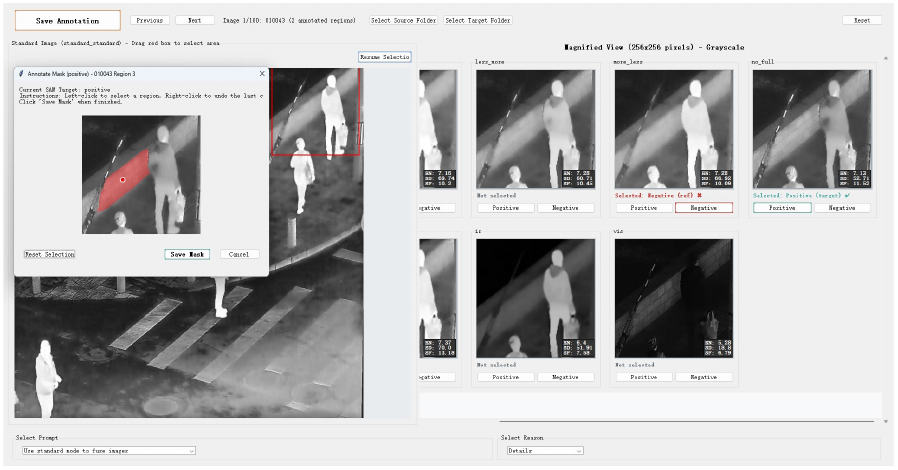}
	\caption{User interface for human feedback annotation. The left side of the interface displays the entire fused image for the user to select a region of interest. The right side shows the fusion results of other candidates within that region, allowing for the annotation of preferred and rejected samples. The pop-up window functions as a SAM-based interface where the user clicks to select the preference region.}
        \label{supp:UI}
        \vspace{-3mm}
\end{figure*}

\noindent\textbf{Human feedback.}
To capture subjective visual quality, we design a custom user interface  (Figure~\ref{supp:UI}) that facilitates the collection of human preferences. Human annotators are presented with candidate fused images generated by PALDM and are asked to select regions according to their perceptual preferences. The segment anything model (SAM)~\cite{kirillov2023segment, SAM} is then applied to accurately generate the corresponding preference-region mask $I_m$ for the selected areas. Specifically, three annotators annotate the feedback datasets for LLVIP, MSRS, and RoadScene, respectively, and we train distinct RLHF models for each dataset.

\noindent\textbf{VLM-based feedback.}
We employ QWEN3-Omni-Think~\cite{xu2025qwen3, QWEN} as a VLM-based feedback mechanism to rank all candidates based on overall image quality and fidelity. We construct three VLM-based feedback datasets for LLVIP, MSRS, and RoadScene, respectively, and train separate RLVF models for each. To ensure the VLM focuses strictly on technical fusion quality rather than high-level semantic content or artistic aesthetics, the model is instructed to act as an ``Image Quality Rater''. The step-by-step instructions are as follows.

\begin{tcolorbox}[colback=gray!5,colframe=gray!40,title=\textbf{ Instruction}]
\begin{itemize}
    \item  You will be given {five} images labeled 1--5. Assess {only perceptual image quality}.
    \item  Focus solely on \textit{signal-level quality}: sharpness, detail retention, structure fidelity, naturalness, and visible artifacts (noise, halos, ringing, blocking, exposure, color cast).
    \item  Ignore content and aesthetics.
    \item  Provide ranking from best to worst using ``$>$'' as separators (e.g., \texttt{3 > 1 > 5 > 2 > 4}).
\end{itemize}
\end{tcolorbox}

Based on the generated ranking, we designate the highest-ranked image as the preferred sample $I_0^w$ and the lowest-ranked image as the rejected sample $I_0^l$. In this scenario, the preference is considered global, and the preference mask $I_m$ is set to encompass the entire image.

\noindent\textbf{Task-driven feedback.}
Downstream evaluations are conducted to examine whether the preference outputs produced by PCLDM align with objective performance metrics.

For the semantic segmentation task, we first process the fused candidate images with SegFormer~\cite{xie2021segformer, segformer} and compare the resulting predictions with the ground-truth labels. The preferred sample $I_0^w$ and the rejected sample $I_0^l$ are determined by evaluating a weighted combination of the mean Intersection over Union (mIoU) and per-class accuracy. The final preference-region mask $I_m$ is defined as the intersection of the predicted masks from the preferred sample, the rejected sample, a baseline reference sample, and the ground-truth mask.

For objective detection, the fused images are sent into a YOLOv11~\cite{khanam2024yolov11, YOLO} to obtain detection results. We generate preferences based on detection performance by computing a weighted sum of mean Average Precision (mAP) and accuracy, which determine the preferred $I_0^w$ and rejected $I_0^l$ samples. We select image patches containing detection targets to serve as the preference regions for IDPO, and the preference mask $I_m$ is set to cover these selected patches.

\begin{figure*}[!t]
	\centering
	\includegraphics*[width=1\linewidth]{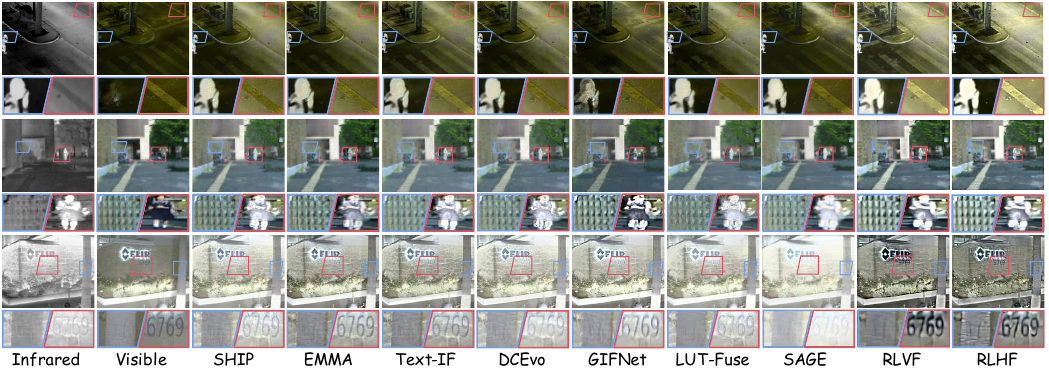}
	\caption{Qualitative comparison of DPOFusion  against state-of-the-art fusion methods on the LLVIP, MSRS, and RoadScene datasets.}
        \label{supp：visual}
\end{figure*}

\begin{table*}[htbp]
\footnotesize
  \centering
  \caption{Quantitative comparison of our methods against state-of-the-art fusion methods on the LLVIP, MSRS, and RoadScene datasets. The best values are in \cellcolor{Best}\textbf{bold}, and the second-best values are \cellcolor{Secondbest}\underline{underlined}.}
  \renewcommand{\arraystretch}{1.1} 
\setlength{\tabcolsep}{1.5mm}{
       \begin{tabular}{ccccccccccccccccc}
    \hline
    \multicolumn{1}{c}{\multirow{2}[1]{*}{Methods}} & \multicolumn{1}{c}{\multirow{2}[1]{*}{Reference}} & \multicolumn{5}{c}{\textbf{LLVIP Dataset}}    & \multicolumn{5}{c}{\textbf{MSRS Dataset}}     & \multicolumn{5}{c}{\textbf{RoadScene Dataset}} \\
    \cmidrule(lr){3-7} \cmidrule(lr){8-12} \cmidrule(lr){13-17} 
          &       & VIF & SCD & SF & DF & VIFF & VIF & SCD & SF & DF & VIFF & VIF & SCD & SF & DF & VIFF \\
          \hline
       U2Fusion & TPAMI$^{22}$ & 0.340  & 1.315  & 11.574  & 4.287  & 0.435  & 0.248  & 1.198  & 7.235  & 3.099  & 0.425  & 0.338  & 1.451  & 12.358  & 6.051  & 0.444  \\
    DDFM  & ICCV$^{23}$ & 0.376  & 1.399  & 13.398  & \cellcolor{Secondbest}\underline{6.785} & 0.428  & 0.369  & 1.387  & 8.856  & 4.738  & 0.482  & 0.360  & \cellcolor{Secondbest}\underline{1.696} & 13.115  & 7.075  & 0.511  \\
    SHIP  & CVPR$^{24}$ & 0.461  & 1.441  & 16.751  & 6.202  & 0.583  & 0.427  & 1.513  & 11.827  & 4.670  & 0.702  & \cellcolor{Secondbest}\underline{0.422} & 1.302  & 15.548  & 7.350  & 0.367  \\
    EMMA  & CVPR$^{24}$ & 0.467  & 1.584  & 14.825  & 5.444  & 0.591  & \cellcolor{Best}\textbf{0.523 } & \cellcolor{Secondbest}\underline{1.629} & 11.559  & 4.387  & 0.773  & \cellcolor{Best}\textbf{0.430} & 1.629  & 15.744  & 7.161  & 0.568  \\
    Text-IF & CVPR$^{24}$ & \cellcolor{Secondbest}\underline{0.491} & \cellcolor{Best}\textbf{1.641} & 15.154  & 5.579  & 0.628  & 0.506  & 1.395  & 10.831  & 4.469  & 0.757  & 0.390  & 1.594  & 16.401  & \cellcolor{Secondbest}\underline{7.692} & 0.538  \\
    DCEvo & CVPR$^{25}$ & 0.484  & 1.546  & 15.882  & 5.341  & 0.599  & \cellcolor{Secondbest}\underline{0.514} & \cellcolor{Best}\textbf{1.665} & 11.460  & 4.326  & 0.756  & 0.410  & 1.476  & 13.860  & 6.246  & 0.442  \\
    GIFNet & CVPR$^{25}$ & 0.356  & 1.505  & \cellcolor{Secondbest}\underline{20.188} & 6.378  & 0.486  & 0.324  & 1.411  & 12.748  & 3.808  & 0.531  & 0.346  & \cellcolor{Best}\textbf{1.728} & \cellcolor{Secondbest}\underline{17.882} & 7.452  & \cellcolor{Secondbest}\underline{0.580} \\
    LUT-Fuse & ICCV$^{25}$ & 0.464  & 1.466  & 14.584  & 4.891  & 0.502  & 0.505  & 1.590  & 11.661  & 4.461  & 0.720  & 0.409  & 1.313  & 12.515  & 5.393  & 0.327  \\
    SAGE & CVPR$^{25}$ & 0.365  & 1.488  & 13.857  & 4.821  & 0.482  & 0.346  & 1.417  & 10.420  & 3.682  & 0.590  & 0.352  & 1.528  & 10.206  & 4.527  & 0.387  \\
    \textbf{RLVF} & Ours  & 0.453  & 1.460  & 18.319  & 6.307  & \cellcolor{Secondbest}\underline{0.803} & 0.459  & 1.456  & \cellcolor{Best}\textbf{17.512} & \cellcolor{Best}\textbf{6.494} & \cellcolor{Best}\textbf{1.134} & 0.361  & 1.475  & \cellcolor{Best}\textbf{23.165} & \cellcolor{Best}\textbf{9.601} & \cellcolor{Best}\textbf{0.625} \\
   \textbf{RLHF} & Ours  & \cellcolor{Best}\textbf{0.495} & \cellcolor{Secondbest}\underline{1.603} & \cellcolor{Best}\textbf{20.733} & \cellcolor{Best}\textbf{6.959} & \cellcolor{Best}\textbf{0.914} & 0.448  & 1.414  & \cellcolor{Secondbest}\underline{17.070} & \cellcolor{Secondbest}\underline{6.269} & \cellcolor{Secondbest}\underline{1.045} & 0.351  & 1.098  & 15.818  & 6.668  & 0.415  \\
    \hline
    \end{tabular}}
  \label{supp：compare_metric}%
\end{table*}%

\section{Additional experimental results}
This section presents additional qualitative results and introduces five other evaluation metrics to comprehensively assess the performance of DPOFusion.
\subsection{Qualitative comparisons}
The extended qualitative comparison results are presented in Figure~\ref{supp：visual}. On the RoadScene dataset, the fusion results of RLVF exhibit richer texture details, with the digital logo on the wall appearing significantly clearer. For the LLVIP dataset, the RLHF fusion results preserve complete traffic markings, aligning closely with human visual perception. Compared to the baseline methods, the fusion results of both RLHF and RLVF meet human perception requirements, offering higher contrast and richer texture details.
\subsection{Quantitative comparisons}
Table~\ref{supp：compare_metric} presents the quantitative comparison results of DPOFusion with other methods on five additional metrics. RLVF and RLHF achieve optimal or comparable values across all three datasets. For LLVIP, RLHF obtains four best values and one second-best value, indicating that the fusion results of RLHF exhibit higher contrast and richer texture details. Meanwhile, RLVF consistently achieves the best or second-best values on the VIFF metric across the three datasets, demonstrating that RLVF effectively fine-tunes the model to generate higher-quality fused images.

\section{Experimental evaluation and analysis}
\begin{figure*}[!t]
	\centering
	\includegraphics*[width=1\linewidth]{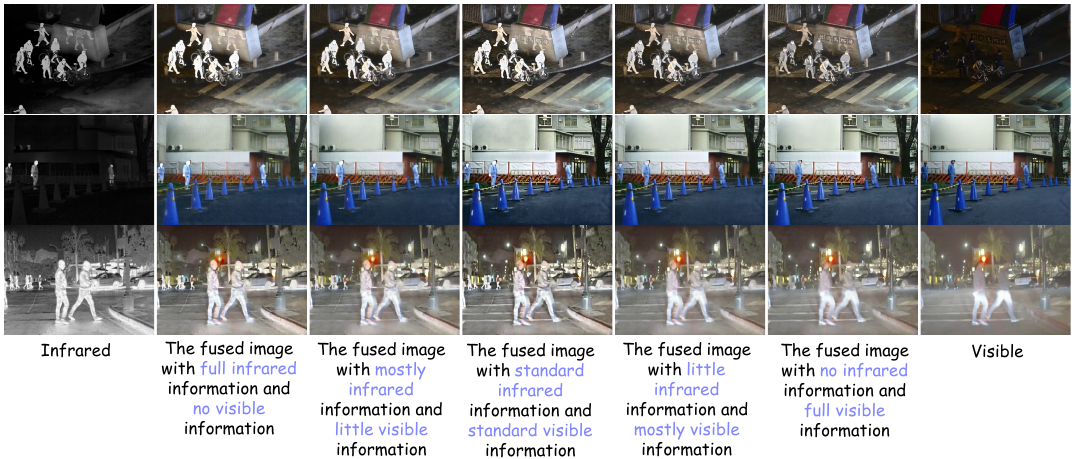}
	\caption{ PALDM results conditioned on corresponding text prompts.}
        \label{ann}
\end{figure*}

\begin{figure}[!t]
	\centering
	\includegraphics*[width=1\linewidth]{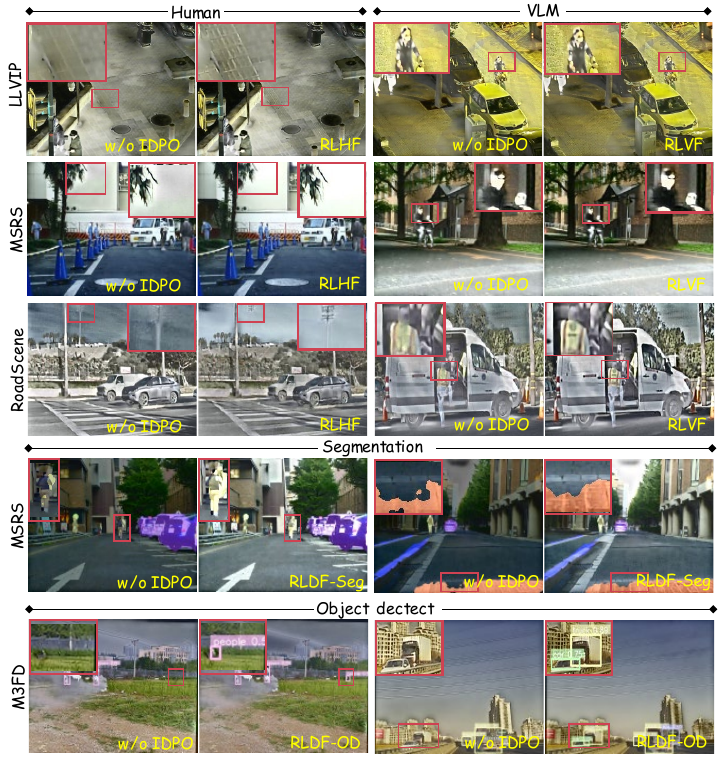}
	\caption{Qualitative results of the IDPO ablation study for DPOFusion.}
        \label{supp:ab_cmpare}
\end{figure}
This section discusses the effectiveness of IDPO on RLHF, RLVF, and RLDF.

\subsection{Fusion results with various text prompts}
To verify the controllability of the property-aligned latent diffusion model (PALDM), we visualize the fusion results generated under different property-descriptive text prompts in Figure~\ref{ann}. By leveraging the proposed joint conditional loss and latent space interpolation strategy, PALDM can flexibly adjust the ratio of modal information in the fused image. In this work, we specifically set the interpolation level to $N=5$. This design choice aims to improve selection efficiency while constraining the solution space distribution of the LDMs. Consequently, it enables the model to more rapidly identify and select appropriate feature mappings within the existing solution space during the subsequent fine-tuning phase. As illustrated in Figure~\ref{ann}, the results exhibit a smooth semantic transition from full infrared information to full visible information. Specifically, images generated with mostly infrared prompts effectively highlight thermal targets, whereas those with mostly visible prompts retain richer background texture details. The diversity results ensure a comprehensive candidate set for the PCLDM fine-tuning.

\begin{table}[htbp]
\small
  \centering
  \caption{Ablation study of the IDPO effectiveness. The best values are shown in \textbf{bold}.}
          \renewcommand{\arraystretch}{1.1} 
\setlength{\tabcolsep}{2.5mm}{
    \begin{tabular}{cccccc}
    \hline
    \multicolumn{6}{c}{LLVIP dataset} \\
    \hline
    Method & EN & SD & AG & MUS & CNN \\
    \hline
    w/o $\mathcal{L}_\text{IDPO}$ & 7.680  & 60.806  & 5.343  & 56.154  & 0.659  \\
                          \cdashline{1-6}
    \rowcolor{gray!10}
    RLVF  & 7.554  & 53.299  & 5.439  & 56.164  & 0.660  \\
     \rowcolor{gray!10}
    RLHF  & \textbf{7.725}  & \textbf{61.911}  & \textbf{5.977}  & \textbf{57.280}  & \textbf{0.660 } \\
    \hline
    \multicolumn{6}{c}{MSRS dataset} \\
    \hline
    Method & EN & SD & AG & MUS & CNN \\
    \hline
    w/o $\mathcal{L}_\text{IDPO}$ & \textbf{7.244}  & 55.158  & \textbf{5.833}  & \textbf{39.707}  & 0.532  \\
                              \cdashline{1-6}
    \rowcolor{gray!10}
    RLVF  & 7.203  & \textbf{56.614}  & 5.782  & 39.684  & 0.524  \\
        \rowcolor{gray!10}
    RLHF  & 7.138  & 53.385  & 5.601  & 39.140  & \textbf{0.545}  \\
    \hline
    \multicolumn{6}{c}{RoadScene dataset} \\
    \hline
    Method & EN & SD & AG & MUS & CNN \\
    \hline
    w/o $\mathcal{L}_\text{IDPO}$ & 7.348  & 45.137  & 6.587  & 43.564  & 0.492  \\
\cdashline{1-6}
    \rowcolor{gray!10}
    RLVF  & \textbf{7.574}  & \textbf{53.323}  & \textbf{8.622}  & \textbf{43.785}  & \textbf{0.539}  \\
        \rowcolor{gray!10}
    RLHF  & 7.210  & 40.535  & 5.957  & 41.431  & 0.472  \\
    \hline
    \end{tabular}}
  \label{tab:supp_three}%
\end{table}%

\subsection{Analysis of IDPO efficacy}
The qualitative results of DPOFusion are presented in Figure~\ref{supp:ab_cmpare}. With IDPO, the framework effectively adjusts the fusion results to align with human visual requirements, as evidenced by the richer texture details on the sidewalks in the LLVIP fusion results. IDPO also post-processes the fusion results to align with large model preferences, ensuring that results on the MSRS and RoadScene datasets retain complete task details. {Finally, IDPO effectively fine-tunes the model to enhance performance in downstream tasks, increasing segmentation accuracy for small targets while improving dense object detection performance.}

The quantitative analysis of the ablation study for IDPO on RLHF and RLVF is presented in Table~\ref{tab:supp_three}. The model fine-tuned with IDPO achieved improved or comparable results across all three datasets. The increases in EN, SD, and AG indicate that IDPO can effectively enhance the information content of fused images with human or VLM feedback, while MUSIQ and CNNIQA metrics demonstrate perceptual quality improvement.
\begin{figure}[!t]
	\centering
	\includegraphics*[width=1\linewidth]{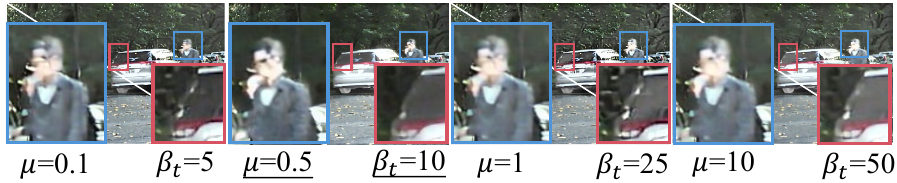}
	\caption{Sensitivity analysis of $\beta_t$ and $\mu$.}
        \label{hepy}
\end{figure}

Table~\ref{tab:supp_od} analyzes the effectiveness of IDPO on object detection with different training strategies. The first two lines of the results present the results of training YOLOv11~\cite{khanam2024yolov11, YOLO} with raw infrared and visible images, as well as the model fine-tuned with PALDM. The remaining few lines display the detection results training with IDPO fine-tuning results for different $\beta_t$. 
Table~\ref{tab:supp_od} indicates that detection accuracy improves as $\beta_t$ increases. This suggests that when utilizing the entire patch as a sample, a larger $\beta_t$ is necessary to effectively constrain the model. We supplement a sensitivity analysis for $\beta_t$ and $\mu$, as shown in Fig.~\ref{hepy}. The settings $\beta_t=10$ and  $\mu=0.5$ yield the clearest human details and the most complete vehicle contours.

\begin{wrapfigure}{r}{0.45\linewidth} 
    \centering
    \includegraphics*[width=\linewidth]{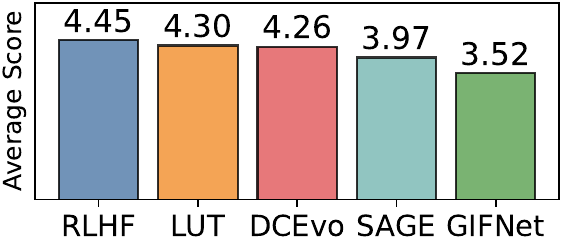}
    \vspace{-7mm} 
    \caption{Preference evaluation result.}
    \label{pref_supp}
\end{wrapfigure}

To verify the fusion quality, as shown in Fig.~\ref{pref_supp},  a blind study with 10 annotators on 20 images uses a 1–5 Likert scale to assess thermal target visibility and texture details (5 = excellent). RLHF exhibits the most prominent thermal targets and texture details.
\begin{table}[!t]
\footnotesize
  \centering
  \caption{Quantitative comparison of the IDPO ablation study for object detection.}
            \renewcommand{\arraystretch}{1.1} 
\setlength{\tabcolsep}{0.4mm}{
    \begin{tabular}{lccccccccc}
    \hline
        Method  & {Bus} & {Car} & {Lamp} & {Moto.} &{People} & {Truck} & @.5:.95 & @.5 & @.75 \\
    \hline
   w/o $\mathcal{L}_\text{IDPO}$ & 74.70  & 62.20  & 33.40  & 34.20  & 38.70  & 56.50  & 49.92  & 76.97  & 52.82  \\
    \cdashline{1-10}
    \rowcolor{gray!10}
    RLDF-OD & \textbf{76.30}  &\textbf{ 63.30}  & \textbf{36.30}  & \textbf{34.80}  & \textbf{41.10}  & \textbf{57.70}  & \textbf{51.58}  & \textbf{78.81}  & \textbf{53.79}  \\
    \hline
    w/o $\mathcal{L}_\text{IDPO}$ & 63.69  & 58.94  & 22.46  & 25.82  & \textbf{39.67}  & \textbf{48.55}  & 43.19  & 66.23  & 45.48  \\
    $\beta_t=10$ & 63.58  & \textbf{58.96}  & 22.69  & 24.69  & 38.88  & 47.86  & 42.77  & 65.20  & 45.93  \\
    $\beta_t=50$ & 63.98  & 58.88  & \textbf{23.49}  & 26.03  & 38.71  & 47.56  & 43.11  & 65.57  & 45.00  \\
        \cdashline{1-10}
    \rowcolor{gray!10}
    $\beta_t=500$ & \textbf{64.07}  & 58.85  & 23.26  & \textbf{26.27}  & 39.45  & 48.50  & \textbf{43.40}  & \textbf{66.48}  & \textbf{46.03}  \\
    \hline
    \end{tabular}}
  \label{tab:supp_od}%
\end{table}%

\subsection{Complexity and runtime comparison}

We evaluate the model size, computational complexity, and inference time of various methods on input images of size $256 \times 256$. Specifically, for the core diffusion components of DPOFusion, the PALDM contains 420.26M parameters with 128.03 GFLOPs, while the PCLDM contains 179.51M parameters with 40.83 GFLOPs.

The DPOFusion performs the diffusion process in a compressed latent space ($64 \times 64$) rather than the original pixel space. Consequently, DPOFusion demonstrates significantly higher efficiency compared to pixel-space diffusion methods like DDFM. As shown in Table~\ref{tab:time}, DDFM requires approximately 7.8s per image, whereas DPOFusion reduces the inference time to 1.7s, achieving substantially fast inference among generative fusion methods.
\begin{table}[!t]
\small
  \centering
  \caption{Comparison of efficiency between DPOFusion and various methods.}
        \renewcommand{\arraystretch}{1.1} 
\setlength{\tabcolsep}{0.9mm}{
    \begin{tabular}{ccccc}
    \hline
    Method & Type  & \multicolumn{1}{l}{Size (M)} & \multicolumn{1}{l}{GFLOPs} & \multicolumn{1}{l}{Time (ms)} \\
    \hline
    U2Fusion & General & 0.66  & 86.44 & 85.51 \\
    SHIP  & General & 0.55  & 35.16 & 33.23 \\
    EMMA  & General & 1.52  & 8.86  & 17.04 \\
            Text-IF & Text-guided & 215.12 & 338.99 & 45.52 \\
    DCEvo & General & 2.02  & 131.36 & 40.10 \\
    GIFNet & General & 0.61  & 39.81 & 13.06 \\
    LUT-Fuse & General & \textbf{0.0078} & - & 2.47 \\

    SAGE  & Semantic-aware & 0.14  & \textbf{4.34}  & \textbf{1.60} \\
    \hline
          \rowcolor{CVPR!8}
          \multicolumn{5}{l}{\textcolor{CVPR}{\textit{Generative methods}}} \\ 
    DDFM  & General & 552.81  & 1840.49 & 7768.12 \\
                      \cdashline{1-4}
    \rowcolor{gray!10}
    DPOFusion & Preference-aligned & 778.4 & 2063.28 & 1709.86 \\
    \hline
    \end{tabular}}
  \label{tab:time}%
\end{table}%

While diffusion-based models generally incur higher computational costs than lightweight CNN-based approaches, DPOFusion is designed primarily as a preference-alignment framework. It serves as a powerful offline generator or teacher model to produce preference-aligned pseudo-labels. These high-quality samples can subsequently be used to supervise or distill lightweight models, allowing the transferred models to achieve real-time inference speeds while inheriting the preference-aligned capabilities of DPOFusion.



\end{document}